\documentclass[11pt,letterpaper]{article}
\pdfoutput=1
\usepackage[nohyperref]{udst2017}
\usepackage{times}
\usepackage[hidelinks]{hyperref}
\usepackage{url}
\usepackage{latexsym}
\usepackage{multirow}
\usepackage{amsmath}
\usepackage{amssymb}
\usepackage{amsfonts}
\usepackage{graphicx}
\usepackage{fancyvrb}

\aclfinalcopy 

\title{A Novel Neural Network Model for Joint POS Tagging and \\ Graph-based Dependency Parsing}

\author{Dat Quoc Nguyen, Mark Dras \and Mark Johnson \\
Department of Computing \\ 
Macquarie University, Australia \\
{\tt {dat.nguyen@students.mq.edu.au}} \\ {\tt \{mark.dras, mark.johnson\}@mq.edu.au}
}

\begin{document}
\maketitle
\begin{abstract}
We present a novel neural network model that learns POS tagging and graph-based dependency parsing jointly. 
Our model uses bidirectional LSTMs to learn  feature representations shared for both POS tagging and   dependency parsing tasks, thus handling the feature-engineering problem. 
Our extensive experiments, on 19 languages from the Universal Dependencies project, show that our model outperforms the state-of-the-art neural network-based Stack-propagation model for joint POS tagging and transition-based dependency parsing, resulting in a new state of the art. 
Our code is open-source and available together
with pre-trained models at: \url{https://github.com/datquocnguyen/jPTDP}.

\medskip

\end{abstract}

\begin{figure*}[t]
\centering
\includegraphics[width=16cm]{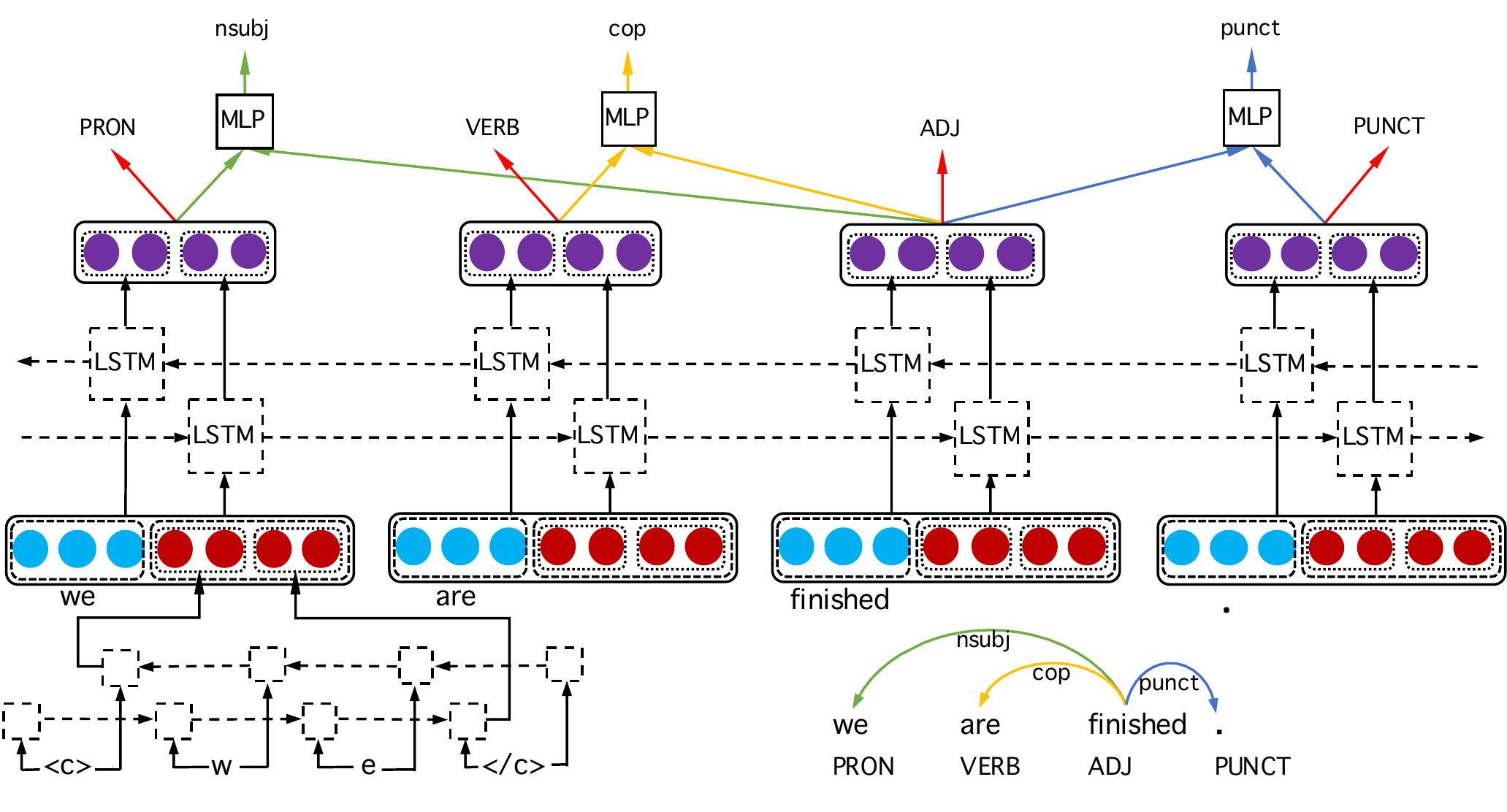}
\caption{Illustration of our jPTDP for joint POS tagging and graph-based dependency parsing.}
\label{fig:model}
\end{figure*}

\section{Introduction}

Dependency parsing has become a key research topic in NLP in the last decade, boosted by the success of the CoNLL
2006, 2007 and 2017 shared tasks on multilingual dependency parsing
\cite{Buchholz2006,Nivre07,udst:overview}. \citet{McDonald2011Nivre} identify two
types of data-driven methodologies for dependency parsing: graph-based approaches
\cite{Eisner:1996,McDonald2005OLT,koo-collins:2010:ACL} and transition-based approaches
\cite{Yamada2003,Nivre2003}.  
 Most traditional graph- or transition-based
parsing approaches 
manually define a set of core and combined features associated with one-hot representations \citep{E06-1011,Nivre2007,Bohnet2010,zhang-nivre:2011:ACL-HLT2011,martinsalmeidasmith2013,choi-mccallum:2013:ACL2013}. Recent work shows that using deep learning in dependency parsing has obtained state-of-the-art performances. Several authors  represent the core features with dense vector embeddings and then feed them as inputs to neural network-based classifiers \citep{chen-manning:2014:EMNLP2014,weiss-EtAl:2015:ACL-IJCNLP,pei-ge-chang:2015:ACL-IJCNLP,andor-EtAl:2016:P16-1}. In addition, others  propose novel neural  architectures for parsing to handle  feature-engineering  \citep{dyer-EtAl:2015:ACL-IJCNLP,cheng-EtAl:2016:EMNLP2016,zhang-zhao-qin:2016:P16-1,wang-chang:2016:P16-1,TACL798,TACL885,DozatM17,MaH17,acl2017dml}.

Part-of-speech (POS) tags are essential features used 
in most dependency parsers. In real-world parsing, those dependency parsers rely heavily on the use of automatically predicted POS tags, thus  encountering  error propagation problems. \citet{Li:2011:JMC:2145432.2145557},  \citet{STRAKA16.873} and \citet{NguyenALTA2016}  show that parsing accuracies  drop  by 5+\%  when utilizing automatic POS tags instead of gold ones. Some attempts have been made to avoid using POS tags during dependency parsing \citep{dyer-EtAl:2015:ACL-IJCNLP,ballesterosEMNLP}, however, these approaches still additionally use the automatic POS tags to achieve the best accuracy. Alternatively, joint learning both POS tagging and dependency parsing has gained more attention because: i) more accurate POS tags could lead to improved parsing performance and ii) 
the the syntactic context of a parse tree could help resolve POS ambiguities \citep{Li:2011:JMC:2145432.2145557,hatori-EtAl:2011:IJCNLP-2011,Lee:2011:DMJ:2002472.2002584,bohnet-nivre:2012:EMNLP-CoNLL,qian-liu:2012:EMNLP-CoNLL,wang-xue:2014:P14-1,zhang-EtAl:2015:NAACL-HLT1,alberti-EtAl:2015:EMNLP,johannsenP16,zhang-weiss:2016:P16-1}.

 In this paper, we propose a novel neural architecture for  joint POS tagging and graph-based dependency parsing. Our model learns latent feature representations shared  for both POS tagging and dependency parsing tasks by using BiLSTM---the   bidirectional LSTM  \cite{Schuster1997BRN,HochreiterSchmidhuber1997b}. 
Not using any external resources such as pre-trained word embeddings,    experimental results on 19 languages from the Universal Dependencies project show that:  our joint model performs better than strong baselines and especially outperforms the 
neural network-based Stack-propagation model for joint POS tagging and transition-based dependency parsing \citep{zhang-weiss:2016:P16-1}, achieving a new state of the art. 

\section{Our joint model}

In this section, we  describe our new model for \textbf{j}oint \textbf{P}OS \textbf{t}agging and  \textbf{d}ependency \textbf{p}arsing, which we call  \textbf{jPTDP}. 
 Figure \ref{fig:model} illustrates the architecture of our new  model.   
We learn shared latent feature vectors representing  word tokens in an input sentence by using BiLSTMs. Then these shared feature vectors are further used to  make the  prediction of POS tags  as well as fed into a multi-layer perceptron with one hidden layer (MLP) to decode  dependency arcs and another MLP to predict  relation types for labeling the predicted arcs.

\paragraph{BiLSTM-based latent feature representations:} 
Given an input sentence $s$ consisting of $n$ word tokens $w_1, w_2, ..., w_n$, we represent each word $w_i$ in $s$ by an embedding $\mathbf{e}^{(\bullet)}_{w_i}$.   \citet{plankP16} and \citet{ballesterosEMNLP} show that character-based  representations of words help improve POS tagging and dependency parsing performances. So, we also use a sequence BiLSTM ($\mathrm{BiLSTM}_{\text{seq}}$) to compute a character-based  vector representation for each word $w_i$ in $s$. 
For a word type $w$ consisting of $k$ characters $w=c_1c_2...c_k$,  the input to the sequence BiLSTM consists of $k$ character embeddings $\mathbf{c}_{1:k}$ in which each embedding vector $\mathbf{c}_{j}$ represents the $j^{\text{th}}$ character $c_j$ in $w$; and  the output is the character-based embedding $\mathbf{e}^{(*)}_w$ of the word type $w$,  computed as:

\setlength{\abovedisplayskip}{-0pt}
\begin{eqnarray*}
\mathbf{e}^{(*)}_w = \mathrm{BiLSTM}_{\text{seq}}(\mathbf{c}_{1:k})
\end{eqnarray*}

For the $i^{\text{th}}$ word $w_i$ in the input sentence $s$, we create an input vector $\mathbf{e}_{i}$ which is a  concatenation ($\circ$) of the corresponding word embedding and character-based embedding vectors:

\begin{eqnarray*}
\mathbf{e}_{i} =  \mathbf{e}^{(\bullet)}_{w_i} \circ \mathbf{e}^{(*)}_{w_i}
\end{eqnarray*}

Then, we feed the sequence of input vectors $\mathbf{e}_{1:n}$ with an additional index $i$ corresponding to a context position  into another  BiLSTM ($\mathrm{BiLSTM}_{\text{ctx}}$), resulting in shared  feature vectors $\boldsymbol{v}_{i}$ representing the $i^{\text{th}}$ words $w_i$ in the  sentence $s$:

\begin{eqnarray*}
\boldsymbol{v}_{i} =   \mathrm{BiLSTM}_{\text{ctx}} (\mathbf{e}_{1:n}, i)
\end{eqnarray*}

\paragraph{POS tagging:} Using shared BiLSTM-based latent feature vector representations, then we follow a common approach to compute the cross-entropy objective loss $\mathcal{L}_{\text{POS}}(\hat{\mathbf{t}}, \mathbf{t})$, in which $\hat{\mathbf{t}}$ and $\mathbf{t}$ are the sequence of predicted POS tags and sequence of gold POS tags of words in the input sentence $s$, respectively \citep{Goldberg16,plankP16}.

\paragraph{Arc-factored graph-based parsing:} Dependency trees can be formalized as directed graphs.  
An arc-factored parsing approach learns the scores of the arcs in a graph \citep{Kubler2009}. Then, using an efficient decoding algorithm (in particular, we use the \citet{Eisner:1996}'s   algorithm), we can find a maximum spanning tree---the highest scoring parse tree---of the graph from those arc scores:

\begin{eqnarray*}
\mathrm{score}(s) =   \underset{\hat{y} \in \mathcal{Y}(s)}{\mathrm{argmax}} \sum\limits_{(h, m) \in \hat{y}} \mathrm{score}_{\mathrm{arc}}(h, m)
\end{eqnarray*}

\noindent where $\mathcal{Y}(s)$ is the set of all possible dependency trees for the input sentence $s$ while $\mathrm{score}_{\mathrm{arc}}(h, m)$ measures the score of the arc between the head $h^{\text{th}}$ word and the modifier $m^{\text{th}}$ word in  $s$. Following \citet{TACL885}, we score an arc by using a MLP with one-node output layer ($\mathrm{MLP}_{\mathrm{arc}}$) on top of the $\mathrm{BiLSTM}_{\text{ctx}}$:

  \begin{eqnarray*}
 \mathrm{score}_{\mathrm{arc}}(h, m) = \mathrm{MLP}_{\mathrm{arc}}(\boldsymbol{v}_{h} \circ \boldsymbol{v}_{m})
\end{eqnarray*}

\noindent where $\boldsymbol{v}_{h}$ and $\boldsymbol{v}_{m}$ are the shared BiLSTM-based feature vectors representing the $h^{\text{th}}$  and   $m^{\text{th}}$ words in $s$, respectively.  We then compute a  margin-based hinge loss $\mathcal{L}_{\text{arc}}$ with loss-augmented inference  to maximize the margin between the gold unlabeled parse tree and the highest scoring incorrect tree \citep{TACL885}.

Dependency relation types are  predicted in a similar manner. We use another MLP on top of the $\mathrm{BiLSTM}_{\text{ctx}}$ for predicting relation type of an head-modifier  arc. Here, the number of the nodes in the output layer of this MLP ($\mathrm{MLP}_{\mathrm{rel}}$) is the number of relation types.  Given an arc $(h, m)$, we compute a corresponding  output vector as:

  \begin{eqnarray*}
{\textbf{v}}_{(h,m)} = \mathrm{MLP}_{\mathrm{rel}}(\boldsymbol{v}_{h} \circ \boldsymbol{v}_{m}) 
\end{eqnarray*}

Then, based on MLP output vectors ${\textbf{v}}_{(h,m)}$, we also compute another margin-based hinge loss $\mathcal{L}_{\text{rel}}$  for relation type prediction, using only the gold labeled parse tree.

\paragraph{Joint model training:} The final training objective function of our joint model is the sum of the POS tagging loss $\mathcal{L}_{\text{POS}}$, the structure loss $\mathcal{L}_{\text{arc}}$ and the relation labeling loss $\mathcal{L}_{\text{rel}}$. The model parameters, including word embeddings, character embeddings,  two BiLSTMs and two MLPs, are learned to minimize the sum of the losses.

\paragraph{Discussion:} Prior neural network-based joint models for  POS tagging and dependency parsing are feed-forward network- and transition-based approaches \citep{alberti-EtAl:2015:EMNLP,zhang-weiss:2016:P16-1}, while our model is a BiLSTM- and graph-based method.  Our model  can be considered as a two-component mixture of a tagging  component and a parsing component. Here, the tagging component can be viewed as a simplified version without  the additional auxiliary loss for rare words of the  BiLSTM-based POS tagging model proposed by \citet{plankP16}. The parsing component can be viewed as an extension of the  graph-based dependency model proposed by \citet{TACL885}, where we replace the input POS tag embeddings by the character-based representations of words.

\begin{table*}[t]
\centering
\resizebox{16.0cm}{!}{
\def\arraystretch{1.05}
  \setlength\tabcolsep{2pt}
  \begin{tabular}{l|lllllllllllllllllll|l}
    \hline
    \multirow{2}{*}{Method} & ar & bg & da & de$^\bullet$ & en & es & eu$^\bullet$ & fa & fi$^\bullet$ & fr & hi & id & it & iw & nl & no & pl$^\bullet$ & pt & sl$^\bullet$ & AVG \\
    \cline{2-21}
    & 10.3 & 12.3 & 15.6 & 11.9 & 9.1 & 7.3 & 17.8 & 8.2 & 24.4 & 5.7 & 4.6 & 13.8 & 5.7 & 10.9 & 18.8 & 11.2 & 23.1 & 10.0 & 19.9 & 12.7\\
    \hline
    \multicolumn{21}{l}{\textsc{Part-of-speech tagging}} \\
    ~~UDPipe & 
    98.7 & 97.8 & 95.8 & 90.7 & 94.5 & 95.0 & 93.1 & 96.9 & \textbf{94.9} & 95.9 & 95.8 & \textbf{93.6} & 97.2 & 94.8 & 89.2 & 97.2 & 96.0 & 97.4 & 95.6 & 95.3 \\
    ~~TnT [$\oplus$] & 
    97.8 & 96.8 & 94.3 & 92.6 & 92.7 & 94.6 & 93.4 & 96.0 & 93.6 & 94.5 & 94.5 & {93.2} & 96.2 & 93.7 & 88.5 & 96.3 & 95.6 & 96.3 & 94.9 & 94.5 \\
    ~~CRF [$\oplus$] & 
    97.6 & 96.4 & 93.8 & 91.4 & 93.4 & 94.2 & 91.6 & 95.7 & 90.3 & 95.1 & 96.0 &  93.0 & 96.4 & 93.6 & 90.0 & 96.2 & 94.0 & 96.3 & 94.8 & 94.2 \\
    ~~BiLSTM-aux & 
    \textbf{98.9} & \textbf{98.0} & \textbf{96.2} & 92.6 & 94.5 & 95.1 & \textbf{94.7} & \textbf{97.2} & \textbf{94.9} & 95.8 & 96.2 &  {93.1} & \textbf{97.6} & \textbf{95.8} & \textbf{93.3} & \textbf{97.6} & \textbf{96.4} & \textbf{97.5} & \textbf{97.6} & \textbf{95.9} \\
    ~~Stack-prop &
    - & - & - & - & - & - & - & - & - & - & - & - & - & - & - & - & - & - & - & 95.4\\
    ~~Our \textbf{jPTDP} &
    98.8 & 97.4 & 95.8 & \textbf{92.7} & \textbf{94.7} & \textbf{95.9} & 93.7 & 96.8 & 94.6 & \textbf{96.0} & \textbf{96.4} & {93.1} & 97.5 & 95.5 & 91.4 & 97.4 & 96.3 & \textbf{97.5} & 97.1 & 95.7\\ 
    ~~~~~~~~~$\bigtriangledown_{\text{-Chars}}$ & 
    3.1 & 2.4 & 3.9 & 2.3 & 1.6 & 0.8 & 4.3 & 0.8 & 5.4 & 1.1 & 0.3 & 3.7 & 1.4 & 1.6 & 6.6 & 2.7 & 4.7 & 3.1 & 5.7 & 2.9\\
    \hline
    \hline
    \multicolumn{21}{l}{\textsc{Dependency parsing}} \\
    ~~UDPipe & 
    76.0 & \textbf{84.7} & 74.8 & 71.8 & 80.2 & 79.7 & 69.7 & 79.7 & \textbf{76.3} & 77.8 & 87.5 & 73.9 & 85.7 & 77.1 & 71.3 & 84.5 & 79.4 & \textbf{81.3} & 80.2 & 78.5 \\
    ~~B'15 [*] &
    75.6 & {83.1} & 69.6 & {72.4} & 77.9 & 78.5 & 67.5 & 74.7 & {73.2} & 77.4 & 85.9 & {72.3} & {84.1} & 73.1 & 69.5 & 82.4 & {78.0} & {79.9} & {80.1} & 76.6 \\
    ~~Pipeline$\mathsf{P}_{tag }$[*]&
    73.7 & 83.6 & 72.0 & 73.0 & 79.3 & 79.5 & 63.0 & 78.0 & 66.9 & 78.5 & 87.8 & 73.5 & 84.2 & 75.4 & 70.3 & 83.6 & 73.4 & 79.5 & 79.4 & 76.6 \\
    ~~RBGParser [*] &
    75.8 & 83.6 & {73.9} & 73.5 & 79.9 & 79.6 & 68.0 & {78.5} & 65.4 & 78.9 & 87.7 & {74.2} & 84.7 & {77.6} & 72.4 & 83.9 & 75.4 & {\bf 81.3} & 80.7 & 77.6 \\      
    ~~Stack-prop &
    {77.0} & {84.3} & 73.8 & {74.2} & {80.7} & {80.7} & {70.1} & {78.5} & {74.5} & {\bf 80.0} & {\bf 88.9} & 74.1 & {85.8} & 77.5 & {\bf 73.6} & {84.7} & {79.2} & 80.4 & {\bf 81.8} & {78.9} \\
    ~~Our \textbf{jPTDP} &
    \textbf{79.0} & 83.9 & \textbf{75.8} & \textbf{75.8} & \textbf{82.0} & \textbf{82.4} & \textbf{73.2} & \textbf{81.5} & {75.0} & \textbf{80.0} & 87.3 & \textbf{75.7} & \textbf{86.4} & \textbf{79.2} & {66.8} & \textbf{84.9} & \textbf{82.5} & 79.3 & 81.7 & \textbf{79.6}\\
    ~~~~~~~~~$\bigtriangledown_{\text{-Chars}}$ & 
    3.8 & 4.1 & 4.5	 & 3.6 & 1.4 & 2.3 & 12.0 & 1.1 & 11.1 & 	0.2 & 0.3	 & 4.1 & 	1.9	 & 1.9 & 	5.4	 & 2.3	 & 10.6	 & 3.4	 & 9.2 & 4.4\\
    \hline
  \end{tabular}
}
\caption{Universal POS tagging accuracies  and LAS scores computed on all tokens (including punctuation) on test sets for 19 languages in UD v1.2. The language codes with $^\bullet$ refer to morphologically rich languages. Numbers (in the second top row) right below language codes are out-of-vocabulary rates. 
\textbf{UDPipe} is the trainable pipeline for processing CoNLL-U files \citep{STRAKA16.873}. 
 \textbf{TnT} denotes  the second order HMM-based TnT tagger \citep{Brants00}. \textbf{CRF} denotes the Conditional random fields-based tagger, presented in \citet{plank-EtAl:2014:Coling}. \textbf{BiLSTM-aux} refers to the state-of-the-art (SOTA) BiLSTM-based POS tagging model with  an additional auxiliary loss for rare words \protect{\citep{plankP16}}.  Note that the (old) language code for Hebrew ``\textbf{iw}'' is referred to as ``\textbf{he}'' as   in \citet{plankP16}.  \textbf{[$\oplus$]}: Results are reported in \citet{plankP16}.  \textbf{Stack-prop} refers to the SOTA Stack-propagation model for joint POS tagging and transition-based dependency parsing \citep{zhang-weiss:2016:P16-1}.   
\textbf{$\bigtriangledown_{\text{-Chars}}$} denotes the  absolute accuracy decrease of our jPTDP, when the character-based representations of words are not taken into account.
\textbf{B'15} denotes the   character-based stack LSTM model for transition-based dependency  parsing  \citep{ballesterosEMNLP}. \textbf{Pipeline$\mathsf{P}_{tag}$} refers to a greedy version of the approach proposed by \citet{alberti-EtAl:2015:EMNLP}. 
\textbf{RBGParser}  refers to the graph-based  dependency parser with tensor decomposition, presented in \citet{lei-EtAl:2014:P14-1}.  \textbf{[*]}: Results are reported in \citet{zhang-weiss:2016:P16-1}.  }
\label{tab:ud-results}
\end{table*}

\section{Experiments}

\subsection{Experimental setup}

Following \citet{zhang-weiss:2016:P16-1} and \citet{plankP16}, we conduct  multilingual experiments on 19 languages from the Universal Dependencies (UD) treebanks\footnote{\url{http://universaldependencies.org/}}  v1.2 \citep{11234/1-1548}, 
using the universal POS tagset  \citep{PetrovDM12}  instead of the language specific
POS tagset.\footnote{ \citet{zhang-weiss:2016:P16-1}  and \citet{plankP16} experimented on 19 and 22 languages, respectively. For consistency, we use 19 languages as in \citet{zhang-weiss:2016:P16-1}.}
For dependency parsing, the evaluation metric is the labeled attachment score (LAS). LAS  is the percentage of words which are correctly assigned both dependency arc and relation type. 

\subsection{Implementation details}\label{ssec:impl}

Our jPTDP is implemented using  \textsc{dynet} v2.0 \citep{dynet}.\footnote{\url{https://github.com/clab/dynet}} We optimize the objective function using Adam  \citep{KingmaB14} with default \textsc{dynet} parameter settings and no mini-batches. We  use a fixed random seed, and we do not utilize pre-trained embeddings in any experiment. Following \citet{TACL885} and \citet{plankP16}, we apply a word dropout rate of 0.25 and Gaussian noise with $\sigma = 0.2$. 
For training, we run for 30 epochs, and  evaluate the \textit{mixed accuracy} of correctly assigning
POS tag together with dependency arc and relation type on the development set after each training epoch. 
We perform a minimal grid search of hyper-parameters on English. We find that the highest mixed accuracy on the English development set is when using 64-dimensional character embeddings, 128-dimensional word embeddings,   128-dimensional BiLSTM states, 2 BiLSTM layers and  100 hidden nodes in MLPs with one hidden layer.\footnote{On English, carried out on a computer with 2.2 GHz Core i7 processor, jPTDP took  6 hours for training with these hyper-parameters,  and then obtained a joint tagging and parsing speed of 700 words/second.} We then apply those hyper-parameters to all 18 remaining languages. 

\subsection{Main results}

Table \ref{tab:ud-results} compares the POS tagging and dependency parsing results of our model jPTDP with  results reported in prior work,  using the same experimental setup.

Regarding POS tagging,  our joint model jPTDP generally obtains similar POS tagging accuracies  to the   BiLSTM-aux  model \citep{plankP16}. Our model also achieves higher averaged POS tagging accuracy  than the  joint   model Stack-propagation \citep{zhang-weiss:2016:P16-1}. There are slightly higher tagging results obtained by  BiLSTM-aux  when utilizing pre-trained word embeddings for initialization, as presented in \citet{plankP16}. However, for a fair comparison to both Stack-propagation and our jPTDP, we only compare to  the results reported without using the pre-trained word embeddings. 

In terms of dependency parsing, in most cases, our  model jPTDP  outperforms Stack-propagation. It is somewhat unexpected that our  model produces about 7\% absolute lower LAS score than Stack-propagation on Dutch (\textbf{nl}). 
A possible reason is
that the hyper-parameters we selected on English
are not optimal for Dutch. Another reason is due to a large number of non-projective  trees in Dutch test set ($106 / 386 \approx 27.5 \%$), while we use the Eisner's decoding algorithm, producing only  projective trees \citep{Eisner:1996}. Without taking ``nl'' into account, our averaged LAS score over all remaining languages is 1.1\% absolute higher than Stack-propagation's.

One reason for our better LAS is probably because jPTDP uses character-based representations of words, while Stack-propagation uses feature representations for suffixes and prefixes which might not be as useful as character-based representations for capturing unknown words. The last row in  Table \ref{tab:ud-results} shows an absolute LAS improvement of 4.4\% on average  when comparing our jPTDP with its simplified version of not  using character-based representations: specifically, morphologically rich languages get an averaged improvement of 9.3 \%,  vice versa 2.6\% for others.\footnote{To determine a morphologically rich language, we take as a proxy for morphological richness the number of noun cases $>= 4$, with this value obtained from WALS (http://wals.info/) where
available  or Wikipedia otherwise.} 
So, our jPDTP is particularly good for morphologically rich languages, with  1.7\% higher averaged LAS than Stack-propagation over these languages.

\section{MQuni at the CoNLL 2017 shared task}

Our team MQuni participated with jPTDP in  the CoNLL 2017 shared task on multilingual parsing from raw text to universal dependencies \citep{udst:overview}.  
Training data are 60+ universal dependency treebanks for 40+ languages from UD v2.0 \citep{ud20data}. 
We do not use any external resource, and we use a fixed random seed and a fixed set of hyper-parameters as presented in Section \ref{ssec:impl}  for all treebanks.\footnote{Except for the biggest treebank UD\_Czech (cs) consisting of 68K training sentences, due to a limited computation resource, we used 64-dimensional word embeddings and 32-dimensional character embeddings. Then it took 30 hours to complete training process for UD\_Czech.} 
For each treebank, we train a joint model for \textit{universal} POS tagging and dependency parsing. We evaluate the mixed accuracy on the development set after each training epoch, and select the model with the highest mixed accuracy. Note that for each ``surprise'' language where there are only few sample sentences with gold-standard annotation or a ``small'' treebank   whose development set is not available, we simply split its sample or training set into two parts with a ratio 4:1, and then use the larger part for training and the smaller part for development.

For parsing from raw text to universal dependencies, we utilize CoNLL-U test files pre-processed by the baseline UDPipe 1.1 \citep{STRAKA16.873}. These  pre-processed  CoNLL-U test files are available to all participants who do not want to train their own models for any steps preceding the dependency analysis, including: tokenization, word segmentation, sentence segmentation, POS tagging and morphological analysis. Note that we only employ the tokenization, word and sentence segmentation, and we do not care about  the POS tagging and morphological analysis pre-processed by  UDPipe 1.1. Recall that we perform universal POS tagging and dependency parsing jointly. In addition, when we encounter an additional parallel test set in a language where multiple training treebanks exist, i.e. a parallel test set marked with language code suffix ``\_pud'' such as ``ar\_pud'', ``cs\_pud'' and ``de\_pud'', we simply use the model trained for its corresponding language code prefix, e.g., ``ar'', ``cs'' and ``de''.

Table \ref{tab:results} presents our official parsing results from the CoNLL 2017 shared task on UD parsing \citep{udst:overview}. We obtain 1\% absolute higher averaged scores than the baseline UDPipe 1.1 \citep{STRAKA16.873} in both categories: big treebank test sets (denoted as \textbf{Big} in Table \ref{tab:results}) and parallel test sets (denoted as \textbf{PUD} in Table  \ref{tab:results}). Specifically, we obtain a highest rank at \textbf{8}$^{\text{th}}$ place for the \textbf{PUD} category, showing that our parsing model jPTDP is particularly good when it is applied to a real practical application in out-of-domain data. Unlike the baseline UDPipe 1.1 and others, for each surprise language, we simply train a joint model just on the sample data of few sentences with gold-standard annotation provided before the test phase, i.e., we utilize neither external resources nor a cross-lingual technique nor a delexicalized parser. So, it is not surprising that we obtain a very low averaged score over the 4 surprise language test sets. When the 4 surprise language test sets are not taken into account, we obtain a rank in top-10 participating systems.

\begin{table}[!t]
\resizebox{7.75cm}{!}{
\def\arraystretch{1.075}
\setlength\tabcolsep{1.5pt}
\begin{tabular}{l|lllll|c}
\hline
 \multirow{2}{*}{\textbf{System}} & \textbf{All} & \textbf{Big} & \textbf{PUD} & \textbf{Sma.} & \textbf{Sur.} &    \multirow{2}{*}{\textbf{R$_{\text{-S}}$}}  \\
& (81) & (55) & (14) & (8) & (4) \\
\hline
UDPipe 1.2  &  69.52$_8$ &  74.38$_{9}$  &  69.00$_{9}$  &  53.75$_{9}$  &  35.96$_{14}$  &  8 \\
UDPipe 1.1  &  68.35$_{13}$  &  73.04$_{17}$  &  68.33$_{13}$  &  51.80$_{15}$  &  37.07$_{11}$  &  15 \\
{{MQuni}} &  {68.05}$_{14}$  &  {74.03}$_{12}$ &  {69.28}$_{8}$  &  {51.58}$_{17}$  &  {14.48}$_{28}$  &  {10} \\
\hline
\end{tabular}
}
\caption{Official macro-averaged LAS F1 scores of  MQuni  and baselines from the CoNLL 2017 shared task on UD parsing \citep{udst:overview}: \protect{\url{http://universaldependencies.org/conll17/results-las.html}}. 
 ``\textbf{All}'' refers to the averaged score over all 81 test sets, which is used as the main metric for ranking participating systems. \textbf{Big}:    the averaged score over 55/81 test sets whose training treebanks are big and have development data available.  \textbf{PUD}: the averaged score over 14/81 test sets that are additional parallel ones, produced separately and their domain may be different from their training data.  \textbf{Sma.}: the averaged score over 8/81 test sets  whose training treebanks are small, i.e., they lack development data and some of them have very little training data.  \textbf{Sur.}: the averaged score over 4/81 remaining  test sets for surprise languages. Here the \textit{subscript} denotes the official rank out of 33 participating systems.  
 {\textbf{R$_{\text{-S}}$}} is the system rank where the 4 surprise language test sets are not taken into account.}
\label{tab:results}
\end{table} 

In fact, it is hard to make a clear comparison between our jPTDP and the parsing models used in other top participating systems. This is because other systems use various external resources and/or better pre-processing modules and/or construct ensemble models for dependency parsing.\footnote{Combining multiple treebanks available for a language or similar languages to obtain larger training data is also considered as a manner of exploiting external data.} For example, UDPipe 1.2 only extends  the word and sentence segmenters of the baseline UDPipe 1.1. Consequently,  UDPipe 1.2 obtains 0.1\% absolute higher in  the macro-averaged word segmentation score\footnote{Word segmentation results are available at: \\ \url{http://universaldependencies.org/conll17/results-words.html}} and 0.2\% higher in the macro-averaged sentence segmentation score\footnote{Sentence segmentation results are available at: \\  \url{http://universaldependencies.org/conll17/results-sentences.html}} than the baseline UDPipe 1.1, resulting in 1+\% better in the macro-averaged LAS F1 score though they use exactly the same parsing model. 
See \citet{udst:overview} for an overview of the methods, algorithms, resources and software used for all other participating systems.\footnote{Outlined at:  \url{http://universaldependencies.org/conll17/systems-in-a-nutshell.html}} 

It is worth noting that for  universal POS tagging, we obtain a highest rank at \textbf{4}$^{\text{th}}$ place for the \textbf{Big} category (i.e., {4}$^{\text{th}}$ on average over 55 big treebank test sets).\footnote{Universal POS tagging results are available at:\\ \url{http://universaldependencies.org/conll17/results-upos.html}}  In this \textbf{Big} category, we also obtain better rank  than both   UDPipe 1.2 and 1.1.

\section{Conclusion}

In this paper, we describe our novel model  for joint POS tagging and graph-based dependency parsing, using bidirectional LSTM-based feature representations. Experiments on 19 languages from the Universal Dependencies (UD) v1.2 show that our model obtains state-of-the-art results in both POS tagging and dependency parsing. 

With our joint model, we  participated in the CoNLL 2017 shared task on UD parsing \citep{udst:overview}.  Given that we followed a strict closed setting while other top participating systems did not,  we still obtained  very competitive results. So, we believe  our joint model can serve as a new strong baseline for further models in both POS tagging and dependency parsing tasks.  

For future comparison, we provide in Table \ref{tab:ud2.0results} the POS tagging, UAS and LAS accuracies  with respect to gold-standard segmentation on the UD v2.0---CoNLL 2017 shared task test sets \citep{ud20testdata}. 
Our  code is open-source and available at: \url{https://github.com/datquocnguyen/jPTDP}.

\begin{table*}[!t]
\resizebox{16cm}{!}{
\setlength\tabcolsep{4pt}
\begin{tabular}{l|cll||l|cll||l|cll }
\hline
\textbf{ltcode} & UPOS & UAS & LAS & \textbf{ltcode} & UPOS & UAS & LAS & \textbf{ltcode} & UPOS & UAS & LAS   \\
\hline
ar\_pud &  79.34 &  68.78 &  56.81 & 
fr\_partut &  95.34 &  84.75 &  80.68 & 
lv &  90.27 &  69.28 &  61.50 \\ 
ar &  95.18 &  84.16 &  77.82 & 
fr\_pud &  89.85 &  83.50 &  78.14 & 
nl\_lassysmall &  95.82 &  79.74 &  75.29 \\ 
bg &  97.49 &  88.53 &  84.20 & 
fr\_sequoia &  97.27 &  86.00 &  83.25 & 
nl &  91.15 &  78.47 &  71.39 \\ 
bxr &  43.21 &  28.79 &  14.04 & 
fr &  96.70 &  87.69 &  84.51 & 
no\_bokmaal &  97.43 &  88.25 &  85.33 \\ 
ca &  98.10 &  88.62 &  85.59 & 
ga &  88.35 &  73.43 &  62.24 & 
no\_nynorsk &  97.07 &  86.30 &  83.12 \\ 
cs\_cac &  98.53 &  87.52 &  83.47 & 
gl\_treegal &  92.83 &  75.45 &  68.46 & 
pl &  96.18 &  88.60 &  82.70 \\ 
cs\_cltt &  97.20 &  79.61 &  74.84 & 
gl &  96.86 &  83.77 &  80.40 & 
pt\_br &  97.64 &  90.40 &  88.32 \\ 
cs\_pud &  95.96 &  85.26 &  79.83 & 
got &  94.27 &  77.78 &  70.27 & 
pt\_pud &  88.41 &  81.49 &  75.15 \\ 
cs &  98.41 &  88.03 &  84.35 & 
grc\_proiel &  94.73 &  73.25 &  67.34 & 
pt &  96.58 &  87.88 &  84.54 \\ 
cu &  92.81 &  81.96 &  73.22 & 
grc &  86.97 &  54.87 &  47.57 & 
ro &  96.72 &  87.04 &  81.37 \\ 
da &  95.80 &  80.87 &  76.89 & 
he &  95.53 &  86.65 &  80.91 & 
ru\_pud &  86.26 &  78.88 &  70.15 \\ 
de\_pud &  85.62 &  78.34 &  71.34 & 
hi\_pud &  85.19 &  64.54 &  51.97 & 
ru\_syntagrus &  98.11 &  89.73 &  87.08 \\ 
de &  92.83 &  80.16 &  75.66 & 
hi &  96.41 &  90.68 &  86.71 & 
ru &  95.31 &  82.14 &  77.12 \\ 
el &  96.18 &  85.07 &  81.55 & 
hr &  96.19 &  85.46 &  79.32 & 
sk &  94.48 &  81.26 &  75.51 \\ 
en\_lines &  94.67 &  79.21 &  74.60 & 
hsb &  51.13 &  29.88 &  17.06 & 
sl\_sst &  88.84 &  63.25 &  55.01 \\ 
en\_partut &  94.17 &  81.25 &  76.56 & 
hu &  91.81 &  74.05 &  66.82 & 
sl &  96.87 &  84.75 &  81.25 \\ 
en\_pud &  94.74 &  85.49 &  81.64 & 
id &  93.10 &  83.41 &  76.84 & 
sme &  33.12 &  22.80 &   8.23 \\ 
en &  94.82 &  85.29 &  81.64 & 
it\_pud &  93.51 &  89.30 &  85.58 & 
sv\_lines &  94.73 &  81.52 &  76.19 \\ 
es\_ancora &  98.28 &  88.48 &  85.50 & 
it &  97.62 &  90.28 &  87.26 & 
sv\_pud &  91.60 &  77.73 &  72.05 \\ 
es\_pud &  88.59 &  87.55 &  80.28 & 
ja\_pud &  97.08 &  94.40 &  93.26 & 
sv &  96.05 &  83.35 &  78.85 \\ 
es &  96.32 &  87.66 &  84.05 & 
ja &  96.56 &  94.07 &  92.41 & 
tr\_pud &  72.60 &  57.14 &  35.50 \\ 
et &  87.62 &  69.44 &  59.15 & 
kk &  51.11 &  44.25 &  22.91 & 
tr &  93.42 &  67.39 &  59.14 \\ 
eu &  93.15 &  77.86 &  72.56 & 
kmr &  47.72 &  31.59 &  18.79 & 
ug &  72.49 &  57.79 &  39.48 \\ 
fa &  96.38 &  85.98 &  81.91 & 
ko &  93.47 &  79.89 &  74.75 & 
uk &  88.09 &  71.03 &  61.03 \\ 
fi\_ftb &  92.63 &  82.48 &  76.54 & 
la\_ittb &  97.44 &  78.81 &  74.65 & 
ur &  92.96 &  86.05 &  79.27 \\ 
fi\_pud &  96.15 &  83.15 &  79.31 & 
la\_proiel &  94.23 &  71.75 &  64.78 & 
vi &  86.78 &  64.88 &  55.63 \\ 
fi &  94.95 &  81.89 &  77.50 & 
la &  83.26 &  57.79 &  44.60 & 
zh &  92.36 &  78.57 &  72.99 \\
\hline
\end{tabular}
}
\caption{Universal POS tagging accuracies (labeled as UPOS), UAS and LAS scores of our jPTDP model with respect to gold-standard segmentation on the UD v2.0---CoNLL 2017 shared task test sets \citep{ud20testdata}. UAS refers to the unlabeled attachment score. \textbf{ltcode} denotes the language treebank code. The  4 surprise language tests are \textit{bxr}, \textit{hsb}, \textit{kmr} and \textit{sme}. The 8 small treebank tests are \textit{fr\_partut}, \textit{ga}, \textit{gl\_treegal}, \textit{kk}, \textit{la}, \textit{sl\_sst}, \textit{ug} and \textit{uk}. The 14 parallel test sets are marked with the language code suffix ``\_pud''. The  55 remaining test sets are for big treebanks.}
\label{tab:ud2.0results}
\end{table*} 

\section*{Acknowledgments} 

This research was supported by a Google award through the Natural  Language Understanding Focused Program, and under the Australian  Research Council's {\em Discovery Projects} funding scheme (project number DP160102156). This research was also supported by NICTA, funded by the Australian Government through the Department of Communications and the Australian Research Council through the ICT Centre of Excellence Program. 
The first author  was supported by  an International Postgraduate Research Scholarship---which is an Australian Government Research Training Program Scholarship---and a NICTA NRPA Top-Up Scholarship.

\bibliographystyle{acl_natbib}
\bibliography{references}

\end{document}